\documentclass[10pt,twocolumn,letterpaper]{article}

\usepackage{wacv}              % To produce the CAMERA-READY version
\usepackage{graphicx}
\usepackage{amsmath}
\usepackage{amssymb}
\usepackage{booktabs}

\usepackage[pagebackref,breaklinks,colorlinks]{hyperref}

\usepackage[capitalize]{cleveref}
\crefname{section}{Sec.}{Secs.}
\Crefname{section}{Section}{Sections}
\Crefname{table}{Table}{Tables}
\crefname{table}{Tab.}{Tabs.}

\usepackage[utf8]{inputenc} % allow utf-8 input
\usepackage[T1]{fontenc}    % use 8-bit T1 fonts
\usepackage{hyperref}       % hyperlinks
\usepackage{url}            % simple URL typesetting
\usepackage{booktabs}       % professional-quality tables
\usepackage{amsfonts}       % blackboard math symbols
\usepackage{nicefrac}       % compact symbols for 1/2, etc.
\usepackage{microtype}      % microtypography
\usepackage{xcolor}         % colors
\usepackage{color}
\usepackage{graphicx}
\usepackage{pifont}
\usepackage{multirow}
\usepackage{xspace}
\usepackage{caption}
\usepackage{amsmath}
\usepackage{enumitem}
\usepackage{wrapfig,lipsum}
\usepackage{diagbox}
\usepackage{makecell}

\newcommand{\cmarkg}{\textcolor{red}{\ding{51}}\xspace}%
\newcommand{\xmarkg}{\textcolor{green}{\ding{55}}\xspace}%
\usepackage{colortbl}

\def\ourmethod{{\textit{MC-TI}}\xspace}
\def\timethod{{\textit{TI}}\xspace}
\def\dicmethod{{\textit{DiC}}\xspace}

\newcommand{\minisection}[1]{\vspace{0.005in} \noindent {\bf #1}}

\newcommand{\tabincell}[2]{\begin{tabular}{@{}#1@{}}#2\end{tabular}}

\newcommand{\model}{\epsilon_\theta}

\newcommand{\conditioner}{\tau}
\newcommand{\expec}{\mathbb{E}}
\newcommand{\encoder}{\mathcal{E}}
\newcommand{\decoder}{\mathcal{D}}

\newcommand{\textprompt}{\mathcal{P}}
\newcommand{\textembedding}{\mathcal{C}}

\newcommand{\tokenset}{\mathcal{V}}
\newcommand{\concept}{\mathcal{V}^{*}}

\newcommand{\pseudoword}{\mathit{S}^{*}}

\newcommand{\inputimage}{\mathcal{I}}

\newcommand{\quotes}[1]{``#1''}
\newcommand{\underit}[1]{\underline{\textit{#1}}}

\def\wacvPaperID{1701} % *** Enter the WACV Paper ID here
\def\confName{WACV}
\def\confYear{2025}

\begin{document}


\title{Multi-Class Textual-Inversion Secretly Yields a Semantic-Agnostic Classifier} 

\author{%
    Kai Wang$^{1}$, Fei Yang$^{3,4}$\thanks{Corresponding author: feiyang@nankai.edu.cn}, Bogdan Raducanu$^{1,2}$, Joost van de Weijer$^{1,2}$\\
    $^1$Computer Vision Center \   $^2$ Universitat Autònoma de Barcelona, Spain\\
    $^3$ VCIP, College of Computer Science, Nankai University, China\\
    $^4$ Nankai International Advanced Research Institute (SHENZHEN· FUTIAN), China\\
    }
  % examples of more authors
  % \And
  % Coauthor \\
  % Affiliation \\
  % Address \\
  % \texttt{email} \\
  % \AND
  % Coauthor \\
  % Affiliation \\
  % Address \\
  % \texttt{email} \\
  % \And
  % Coauthor \\
  % Affiliation \\
  % Address \\
  % \texttt{email} \\
  % \And
  % Coauthor \\
  % Affiliation \\
  % Address \\
  % \texttt{email} \\

% \and

\maketitle

%%%%%%%%% ABSTRACT
\begin{abstract}
With the advent of large pre-trained vision-language models such as CLIP, prompt learning methods aim to enhance the transferability of the CLIP model.
They learn the prompt given few samples from the downstream task given the specific \textit{class names} as prior knowledge, which we term as \textit{semantic-aware} classification.
However, in many realistic scenarios, we only have access to few samples and no knowledge of the class names (e.g., when considering instances of classes). This challenging scenario represents the \textit{semantic-agnostic} discriminative case.
Text-to-Image (T2I) personalization methods
aim to adapt T2I models to unseen concepts by learning new tokens and endowing these tokens with the capability of generating the learned concepts. 
These methods do not require knowledge of class names as a \textit{semantic-aware} prior. Therefore, in this paper, we first explore Textual Inversion and
reveal that the new concept tokens possess both generation and classification capabilities by regarding each category as a single concept. 
However, learning classifiers from single-concept textual inversion is limited since the learned tokens are suboptimal for the discriminative tasks. To mitigate this issue, we propose Multi-Class textual inversion, which includes a discriminative regularization term for the token updating process. 
Using this technique, our method \ourmethod achieves stronger \textit{S}emantic-\textit{A}gnostic \textit{C}lassification while preserving the generation capability of these modifier tokens given only few samples per category. In the experiments, we extensively evaluate \ourmethod on 12 datasets covering various scenarios, which  demonstrates that \ourmethod achieves superior results in terms of both classification and generation outcomes.
\end{abstract}

\begin{figure*}[t]
\centering
\includegraphics[width=0.825\textwidth]{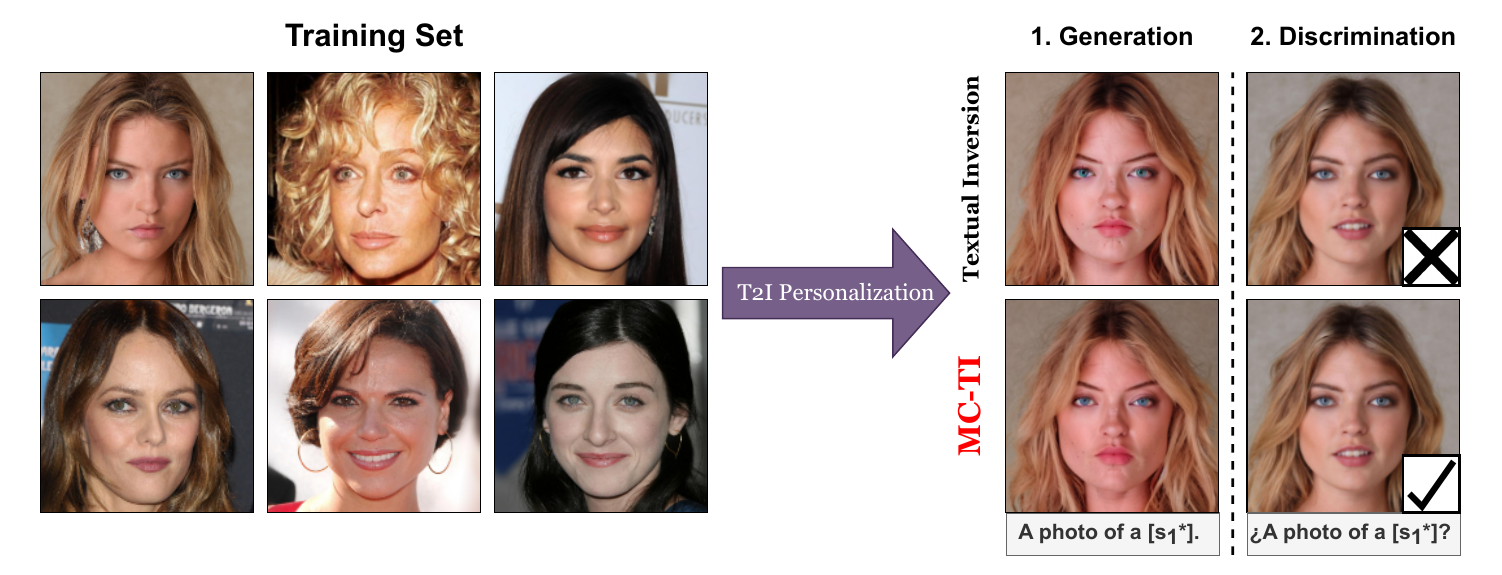}
\vspace{-5mm}
\caption{
While both existing Single-Concept Textual Inversion and our proposed Multi-Class Textual Inversion (\ourmethod) can generate satisfactory results with \textit{few samples} per person, the single-concept \timethod lacks the ability to ensure discrimination performance. This is because it does not constrain token updates in a discriminative manner. (CelebA faces~\cite{liu2015celeba}.)
}
\vspace{-6.5mm}
\label{fig:problem_setup}
\end{figure*}

\section{Introduction}
\label{sec:intro}

Leveraging extensive datasets of image-text pairs, trained visual-language models (VLMs)~\cite{alayrac2022flamingo,jia2021ALIGN,radford2021clip} encapsulate critical general knowledge. Different from traditional representation learning based on discretized labels, the alignment between image and text features endows the VLMs with superior generalization capabilities for downstream tasks. 
While VLMs are effective in extracting both visual and textual descriptions, their training requires large-scale, high-quality datasets. 
To circumvent this issue, prompt learning methods~\cite{zhou2022coop,zhu2023prograd} adapt a pre-trained VLM (e.g., CLIP~\cite{radford2021clip}) to downstream tasks. They have demonstrated impressive performance across a variety of few-shot and zero-shot visual recognition cases. 
Despite their effectiveness, these prompting methods typically require knowledge of \textit{class names} for context optimization, which we term \textit{semantic-aware} classification. This requirement potentially limits the applicability of these methods in realistic scenarios where class names are not determined.

To learn token representations for new concepts, recent T2I personalization methods~\cite{textual_inversion,kumari2022customdiffusion,ruiz2022dreambooth} propose to adapt a given T2I diffusion model with user-provided images and associating the new concept with a unique token as their own \textit{``names''}. Consequently, the adapted model can generate various renditions of the new concept guided by text prompts. 
However, these learned tokens only address generative paradigms. As mentioned in the seminal paper~\cite{hinton2007recognize}, generative approaches are also crucial for discrimination tasks. Therefore, it should be possible to \textit{unify} both discriminative and generative paradigms in the T2I model personalization case, such a problem setup is shown in Fig.~\ref{fig:problem_setup}.
A recent work, named Diffusion Classifier (\dicmethod)~\cite{li2023diffusion_classifier}, first examines how T2I diffusion models compare against discriminative models. Nonetheless, the \dicmethod method  requires knowledge of category names as a \textit{semantic-aware} prior, similar to the prompt learning methods.
Consequently, these methods are not suitable for discrimination tasks involving unknown concepts that are difficult to name before training, which we refer to as the \textit{semantic-agnostic} classification.
Moreover, each unknown concept typically has only \textit{few-shot} samples available, which further complicates the task.

\begin{figure*}[t]
\centering
\includegraphics[width=0.845\textwidth]{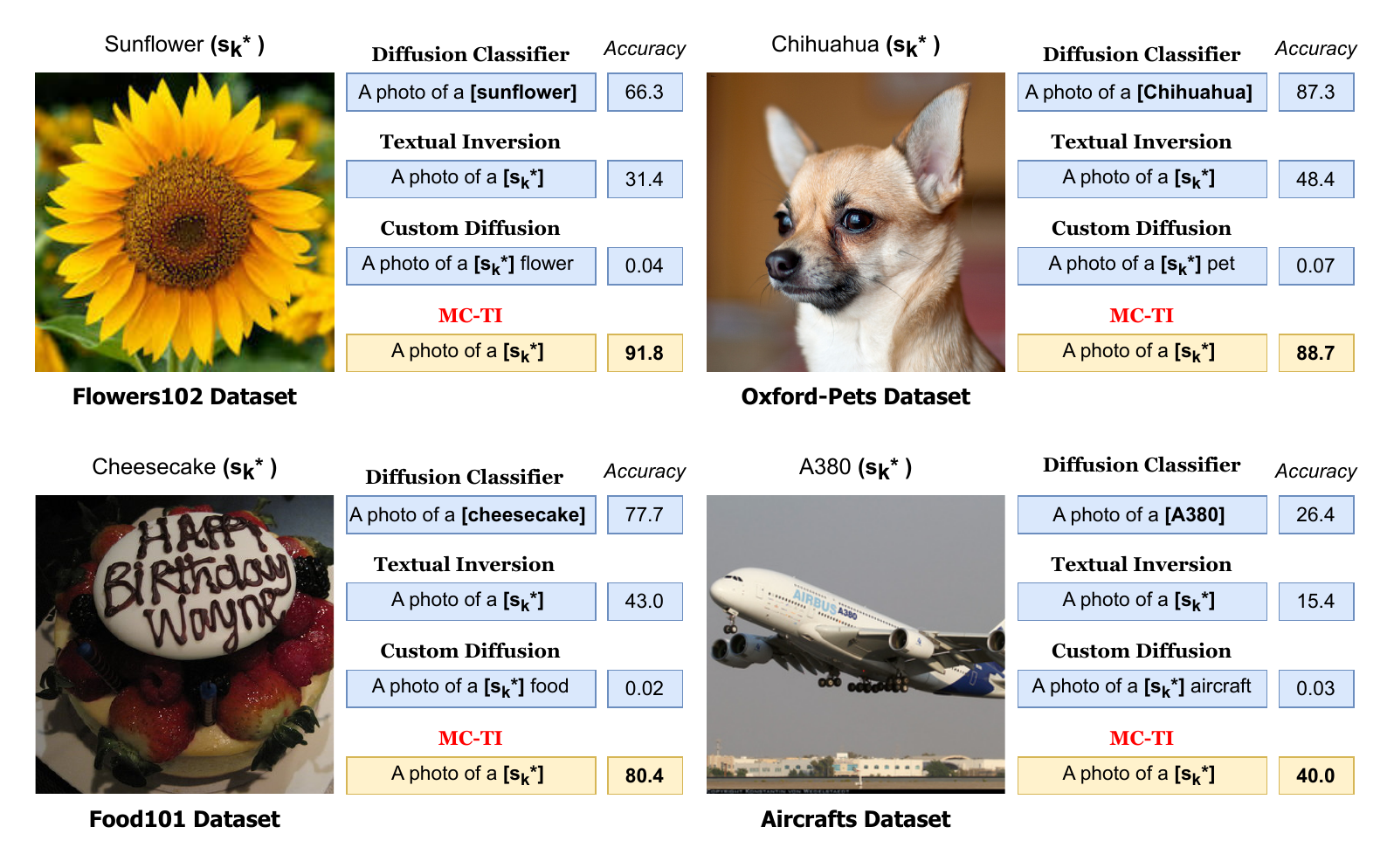}
\vspace{-6mm}
\caption{Diffusion Classifier~\cite{li2023diffusion_classifier} approach classifies samples by computing the text-conditional likelihoods and requires knowledge of the \textit{category names}. 
By comparison, Textual Inversion (\timethod)~\cite{textual_inversion} and \ourmethod only need to learn the concept tokens with few-shot samples (\textit{5-shot} in the figure examples). \ourmethod further strengthen the \timethod by augmenting with a discriminative regularization term and significantly improve the performance.
Custom Diffusion~\cite{kumari2022customdiffusion} is one of the best personalization methods by fine-tuning the UNet, it works poorly in classification. 
}
\vspace{-5mm}
\label{fig:teaser}
\end{figure*}

To achieve \textit{semantic-agnostic} classification with few samples, we first explore a naive scenario where we directly employ tokens learned with Textual Inversion (\timethod) for classification (as shown in Fig.~\ref{fig:teaser} for 5-shot classification), which exhibits much lower accuracies compared with \dicmethod but still demonstrates the potential applicability in the classification task.
Furthermore, observing that Custom Diffusion~\cite{kumari2022customdiffusion}
behaves poorly in the classification task, we hypothesize that this discrepancy is due to interference caused by backbone fine-tuning. In this paper, we build on the frozen backbone method \timethod  as our primary baseline.

On the basis of the above observations, we conclude that the current personalization algorithms primarily emphasize the generation ability of new tokens. More specifically, they update the tokens solely based on the noise reconstruction loss.
While these approaches provide more freedom in updating the tokens, it may compromise their discriminative capabilities (as shown in Fig.~\ref{fig:text_feature}). 
To enhance the training process and guide these tokens to be applicable to the \textit{semantic-agnostic} classification, we propose \emph{Multi-Class textual inversion} (instead of the existing single-concept textual inversion). 
In our method, we propose constraining the updating paths to follow a \textit{discrimination-regularized} direction. 
In other words, the tokens are required to be updated in a manner that leads to a better distribution to achieve discriminative textual prompt representations. 
More specifically, in each training step, we randomly sample image features for each category (concept). These features contribute to composing a temporary classifier for the current textual prompt, which already includes the learnable token. Subsequently, we compute the classification probability between the textual prompt and these image features based on their cosine similarities. To ensure discrimination, a cross-entropy loss is applied to the probability scores, serving as a \textit{discriminative regularization}. This loss is added to the original $\epsilon$-prediction loss from the personalization approaches.

Our method, designated as Multi-Class Textual Inversion (\ourmethod)
is evaluated across twelve recognition datasets covering various recognition scenarios.
It demonstrates remarkable performance in \textit{Semantic-Agnostic Classification} tasks with few-shot samples. 
Notably, even compared to the strong \textit{semantic-aware} zero-shot method, namely CLIP (ViT-L/14)~\cite{radford2021clip}, which rely on class names trained over millions of image-text pairs, \ourmethod outperforms them with as few as 1-shot (Flowers, EuroSAT) and up to 8-shot (Stanford-Cars, Aircrafts, DTD). 
As a summarization, our method has the following main contributions:

\begin{itemize}[leftmargin=*]
    \item We are the \textit{first} paper to explore \textit{semantic-agnostic classification} of T2I personalized tokens. By this means, we unify the generative and discriminative paradigms upon T2I diffusion adaptation.
    \item We propose \textit{Multi-Class textual inversion} that includes a simple but effective discriminative regularization term, which is defined as a cosine cross-entropy loss, to augment the personalization method for discrimination. 
    \item We conducted experiments over twelve benchmarks covering diverse recognition tasks and various $N$-shot settings. These experimental evaluations demonstrate the efficiency of our proposed \ourmethod.
\end{itemize}

\section{Related Work}

\minisection{Generative models for classification}
works~\cite{hinton2007recognize,ng2001discriminative,ranzato2011deep}
have emphasized the importance of modeling data distributions to facilitate discriminative learning. Additionally, in the realm of generative modeling, efforts have been made to acquire efficient representations for classification tasks, as evidenced by works like MAE~\cite{he2022masked}, BERT~\cite{devlin2018bert},  etc. However, these endeavors typically involve joint training for discriminative and generative modeling or fine-tuning generative representations for downstream tasks.
A recent contribution, known as the \dicmethod~\cite{li2023diffusion_classifier}, takes a different approach by directly using the Stable Diffusion~\cite{Rombach_2022_CVPR_stablediffusion} for discrimination tasks. This method computes text-conditional likelihoods by estimating noise reconstruction losses, demonstrating promising performance compared to discriminative approaches in zero-shot scenarios. However, the \dicmethod incurs significantly longer diffusion sampling times for each input image, rendering it impractical for real-time classification with large numbers of samples. Furthermore, it also imposes the prerequisite of knowing the class names as a \textit{semantic-aware} prior.

\minisection{Text-to-Image models personalization} 
aims at adapting a given model to a \textit{new concept} by giving users images and bind the new concept with a unique token. As a result, the adaptation model can generate various renditions of the new concept guided by text prompts. Depending on whether the adaptation method is fine-tuning the T2I model, they are categorized into two main streams: 
The freezing  stream focuses on learning new concept tokens instead of fine-tuning the generative models.
Textual Inversion~\cite{textual_inversion} is a pioneering work focusing on finding new pseudo-words by personalizing the text embedding space. Recent methods~\cite{hiper2023,li2023photomaker,gu2024mixofshow,kai2023DPL,li2023stylediffusion} also belong to this technique stream.
The most representative methods of fine-tuning stream include DreamBooth~\cite{ruiz2022dreambooth} and Custom Diffusion~\cite{kumari2022customdiffusion}, where the pretrained T2I model such that it learns to bind a unique identifier with that specific subject given few images. Following research~\cite{gal2023e4t,han2023svdiff,Cones2023,butt2025colorpeel} further extend this pipeline and improve the generation quality.
Although existing T2I model adaptation methods have been successful in learning new concepts from a set of relevant images, they have overlooked that personalized concept tokens have gained semantic information from the relevant images. These inherent semantic information from these tokens is also applicable to image classification tasks, as shown in Fig.~\ref{fig:teaser}. To avoid misalignment of text and image pairs, we adopt the branch of the T2I model freezing, that is, the Textual Inversion~\cite{textual_inversion} method as the backbone.

\minisection{Prompt Learning } is
to adapt the pretrained Visual-Language Models (VLMs) to the downstream tasks, prompt learning always applies task-related textual tokens to infer the task-specific textual knowledge. For example, the hand-crafted template “a photo of a [CLASS]” in CLIP~\cite{radford2021clip} is used to model the textual embedding for zero-shot prediction with knowing the \quotes{[CLASS]} as the \textit{semantic-aware} prior. However, the hand-crafted prompts have fewer ability to describe the downstream task because they do not consider the specific knowledge of the current task. To address the above problem, Context Optimization (CoOp)~\cite{zhou2022coop} replaces hand-crafted prompts with a soft learning prompt inferred by the labeled few-shot samples.
PLOT~\cite{chen2022plot} proposes to apply optimal transport to match the vision and text modalities.
Following these pioneering works, recent methods~\cite{gao2024clip_adapter,yao2023kgcoop,zhu2023prograd} continue to improve the prompt learning performance.

However, these prompt-tuning methods have the limitation that they require \textit{semantic-aware} prior knowledge of the concept names before training. Furthermore, prompt learning involves the update of the network on \textit{all samples} collected in each training time, and does not support \textit{parallel} training as our method \ourmethod. More importantly, the learned context tokens from these methods cannot be further applied in \textit{image generation}.

\section{Method}
\label{sec:method}

\subsection{Preliminaries}
\label{sec:preliminary}
\minisection{T2I Diffusion Models.}
In this paper, we utilize Stable Diffusion~\cite{Rombach_2022_CVPR_stablediffusion} as our backbone model, which is a latent diffusion model (LDM). 
The model comprises two primary components: an autoencoder and a diffusion model applied in the latent space.
The encoder $\encoder$ within the autoencoder segment of the LDMs maps an input image $\inputimage$ to a latent code $z_0=\encoder(\inputimage)$, while the decoder reverses this process, reconstructing the original image as $\decoder(\encoder(\inputimage)) \approx \inputimage$.
The diffusion model can be conditioned on various factors such as class labels, segmentation masks, or textual input. Let $\tau(y)$ denote the conditioning mechanism, which maps a condition $y$ to a corresponding conditional vector for LDMs. The LDM model is updated using the noise reconstruction loss, also known as the $\epsilon$-prediction loss:
\begin{equation}
L_{LDM} = \expec_{z_0 \sim \encoder(x), y, \epsilon \sim \mathcal{N}(0, 1)}  \; \underbrace{\Vert \epsilon - \model(z_{t},t, \conditioner(y)) \Vert_{2}^{2}}_{\mathcal{L}_{mse}}. 
\label{eq:ldm_loss}
\end{equation}
The neural network backbone $\model$ typically adopts a conditional UNet architecture~\cite{ronneberger2015unet}, responsible for predicting the added noise. In text-guided diffusion models, the objective is to generate an image based on a combination of random noise $z_T$ and a conditional input prompt $\textprompt$. To distinguish from the general conditional notation in LDMs, we represent the textual condition as $\textembedding=\tau_\phi(\textprompt)$.  $\tau_\phi$ refers to a CLIP text encoder pretrained on millions of text-image pairs.

\minisection{T2I model adaptation.}
Given a pretrained T2I diffusion model, adaptation methods~\cite{kumari2022customdiffusion,ruiz2022dreambooth,textual_inversion} integrate a new concept into the model using few images and their associated text descriptions. 
Typically, this involves learning new tokens through text encoding.

\minisection{Single-Concept Textual Inversion}. Given the target concept with few images, a text caption is required. For personalization purpose where the target concept is a unique instance of a general category, we introduce a new modifier token $\concept$ associated with the pseudo-word $\pseudoword$ for the concept.
During training, $\concept$ is initialized with the embeddings of a single-word coarse descriptor of the object and injected to a prompt template of the form \quotes{A photo of a $\pseudoword$}, \quotes{A nice photo of the small $\pseudoword$}, etc.
Then the optimization goal with the LDM loss $L_{TI} = L_{LDM}$ is:
\begin{equation}
\concept = \underset{\tokenset}{\arg\min} \   \expec_{z_0 \sim \encoder(x), y, \epsilon \sim \mathcal{N}(0, 1)} \;  \mathcal{L}_{mse}.  
\label{eq:ti_loss}
\end{equation}
During this optimization, the network parameters are frozen and only the $\concept$ tokens are learned.

\begin{figure*}[t]
\centering
\includegraphics[width=0.85\textwidth]{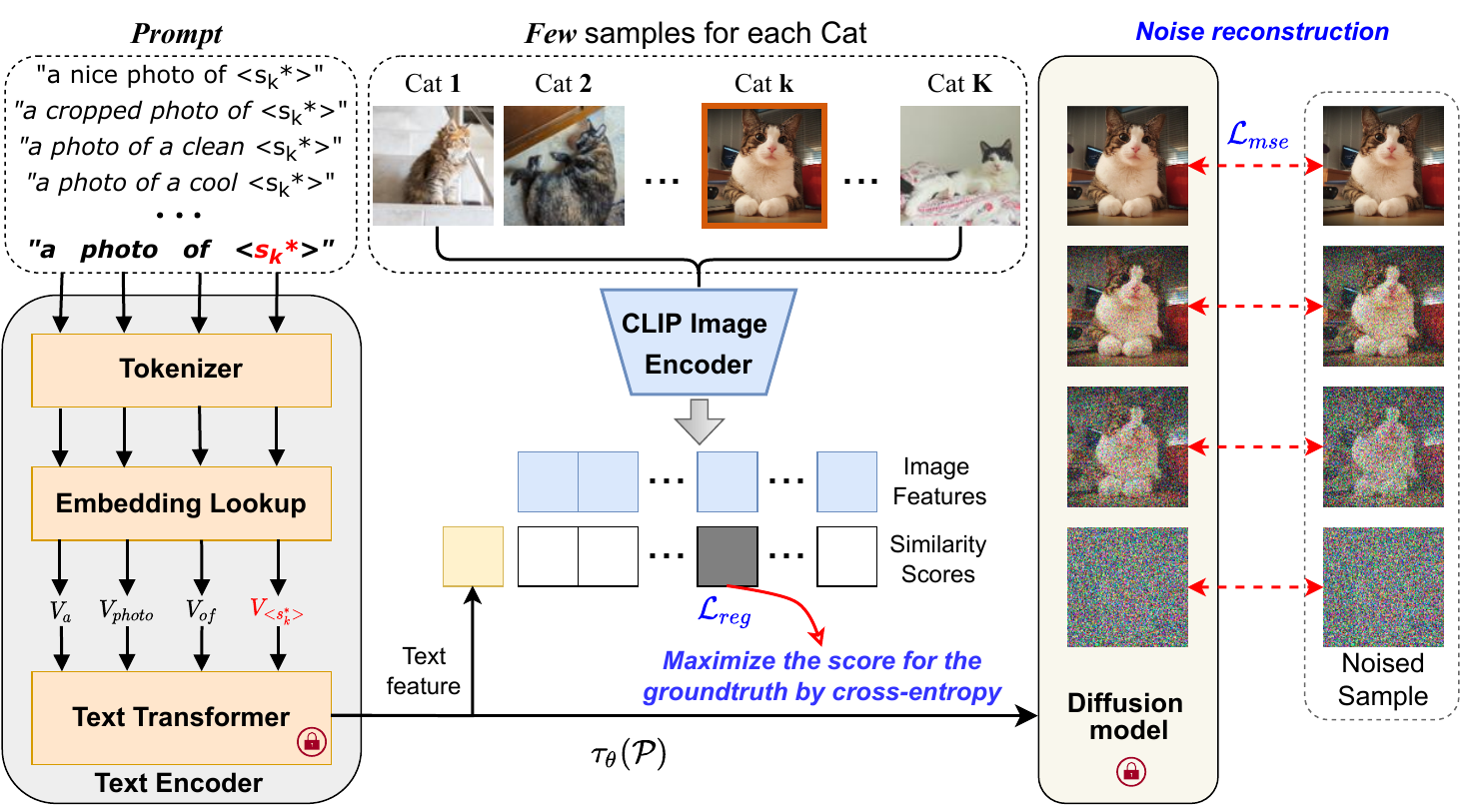}
\vspace{-3mm}
\caption{Illustration of our \ourmethod approach. During the training of token <$s^*_k$> for concept $k$ (cat $k$ for example), we add a discriminative regularization term $\mathcal{L}_{reg}$,
which is defined as the cosine cross-entropy from the current training text feature to the image features.
}
\vspace{-3mm}
\label{fig:method}
\end{figure*}

\subsection{\ourmethod: Multi-Class Textual Inversion}
\label{sec:ourmethod}

\begin{figure*}[t]
\centering
\includegraphics[width=0.9\textwidth]{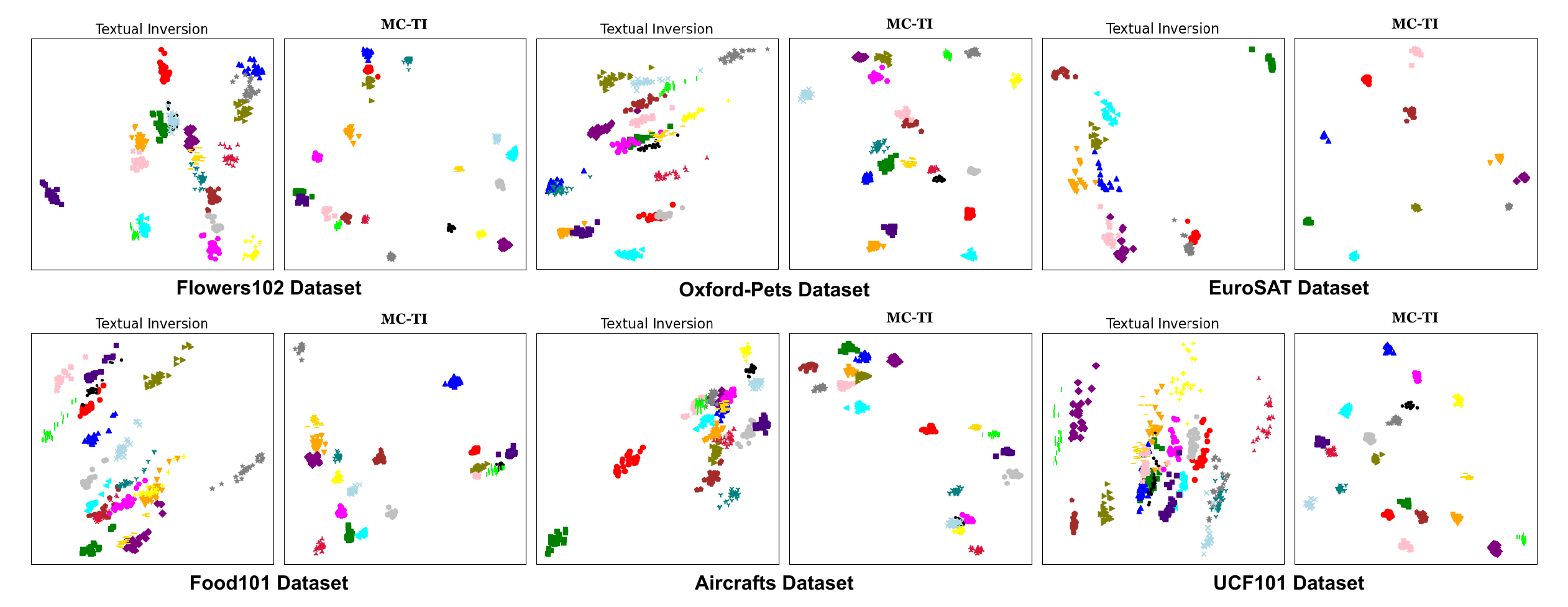}
\vspace{-4mm}
\caption{
To visualize the textual prompts features, we took the 5-shot conceptual tokens learned by Textual Inversion and \ourmethod, respectively. By applying 27 types of various prompt templates, we visualize the PCA components in 2-D maps for 20 categories out of these six datasets. 
\ourmethod improves the clustering of textual characteristics by enforcing discriminative regularization terms.
}
\vspace{-5.5mm}
\label{fig:text_feature}
\end{figure*}

\minisection{Semantic-Agnostic Classification.}
Suppose we already have a dataset with \textit{multiple concepts} as $K$ classes, each class is given with only $N$-shot ($N=1,2,4,5,8,16$) samples for training purposes, which are denoted as $\inputimage_k^n, n \in [1,N], k \in [1,K]$. 
Importantly, we lack \textit{semantic} information about these classes, which means that we do not know their class names in advance. 
This \textit{semantic-agnostic} scenario occurs when the classes we aim to learn do not have semantic labels (or their semantic labels are unknown to the language model, e.g. person names). This is the case with instances of human faces, very specialized fields such as fine-grained class names, etc. Then we aim to learn a specific token $\concept_k$ for each class pseudo-word $\pseudoword_k$ separately, and these tokens $\concept_k$ can help us to discriminate images from the same dataset distributions while maintaining their generation capability.
Training solely with the noise  $\epsilon$-prediction loss in \textit{single-concept} Textual Inversion (\timethod) can only ensure that the learned tokens are able to generate the desired concepts.
However, as the tokens learned from the pretrained diffusion model already implicitly include the semantic information of the given images, they can already achieve rough discrimination tasks (as seen in Fig.~\ref{fig:teaser}).
Nonetheless, the \timethod approach does not constrain the optimization direction of these tokens with discriminative guidance. This results in a lower classification performance than for diffusion-based classifiers, like \dicmethod~\cite{li2023diffusion_classifier}.
The PCA visualization depicted in Fig.~\ref{fig:text_feature} illustrates the textual features learned by the \timethod. In particular, these features often overlap among categories, indicating a lack of distinctiveness. To address this issue and improve classification accuracy, we propose to apply a \textit{discriminative regularization} to achieve \textit{Multi-Class textual inversion} (\ourmethod).

\minisection{Discriminative Regularization.}
For convenience, we denote the CLIP image encoder as $\pi_{\psi}$, which is trained paired as the CLIP text encoder $\tau_{\phi}$. During the token learning for concept $\concept_k$, we first inject $\concept_k$ into a prompt template and obtain its corresponding textual feature as $g_k = \tau_{\phi}(\textprompt^{\concept_k})$ 
we randomly sample \textit{one image} for each class out of the \textit{multiple concepts} in the few-shot training set $\inputimage^n_j, j \in [1,K]$ and pass them to the image encoder for feature extraction as $f_j = \pi_{\psi}(\inputimage^n_j)$. In this way, we have a group of features $f_1,f_2,...,f_j,...,f_K$ representing all $K$ classes. Regarding these features as temporal prototypes, we propose to compute the prediction probability for the textual feature as follows:
\begin{align}
P(\hat{y}=j|g_k) = \frac{s \cdot  exp(cos(g_k,f_j)) }{\sum_{j=1}^K  s \cdot exp(cos(g_k,f_j))}
\label{eq:prob_pred},
\end{align}
where $s$ is the scale factor as the reciprocal of the softmax temperature and $\hat{y}$ is the probability prediction for the textual feature with token $\concept_k$. Then the cross-entropy loss is:
\begin{align}
\mathcal{L}_{reg} = -\sum_{j=1}^K y_j \cdot log P(\hat{y}=j|g_k)
\label{eq:reg_loss}.
\end{align}
Here $y$ denote the one-hot groundtruth.
Finally, the loss for training the token $\concept_k$ is defined as:
\begin{equation}
\concept_k = \underset{\tokenset_k}{\arg\min} \   \expec_{z_0 \sim \encoder(x), y, \epsilon \sim \mathcal{N}(0, 1)} \; \Big[ \alpha \mathcal{L}_{mse} + \beta \cdot \mathcal{L}_{reg} \Big].
\label{eq:final_loss}
\end{equation}
The $\alpha, \beta$ work as the trade-off parameters to balance these two losses.
Note that each token $\concept_k$ is updated separately for each concept $\pseudoword_k$. Therefore, we can obtain a set of $\concept_k, k \in [1,K]$ in \textit{parallel}.
After applying our robust regularization terms, the textual features are better distributed for future discrimination tasks, as shown in Fig.~\ref{fig:text_feature}.

\minisection{Inference.} 
During inference time, each sample $\inputimage$ is classified by predicting its probability of belonging to all the learned \textit{multiple concepts} (tokens).
To achieve this, we first inject all category tokens $\concept_k$ into the prompt template $T=$ \quotes{a photo of a $\pseudoword_k$} and then extract the textual features as $g_k = \tau_{\phi}(\textprompt^{\concept_k}), k \in [1,K]$. The input image with feature as $f_{\inputimage} = \pi_{\psi}(\inputimage)$ is categorized as:
\begin{align}
P(\bar{y}=k|f_{\inputimage}) = \frac{s \cdot  exp(cos(g_k,f_{\inputimage})) }{\sum_{l=1}^K  s \cdot exp(cos(g_l,f_{\inputimage}))}
\label{eq:inference}.
\end{align}
Where the $\bar{y}$ is the prediction probabilities for the input $\inputimage$. 
Moreover, given that \ourmethod is orthogonal to prompt learning methods, we can enhance our classification performance by integrating these approaches. In this study, we specifically employ the CoOp~\cite{zhou2022coop} to learn a unified context.

\section{Experiments}
\label{sec:experiment}

\subsection{Experimental setups}
\minisection{Implementation Details.}
We utilize the Stable Diffusion (SD) v1.5~\cite{Rombach_2022_CVPR_stablediffusion} as our backbone. Within the SD model, we employ the text branch of CLIP (ViT-L/14)~\cite{radford2021clip} as the text encoder $\tau_{\phi}$, and we additionally utilize the image encoder $\pi_{\psi}$ to extract features for the training samples.
We also include SD v1.4 and v2.0 in the ablation study.
For \textit{single-concept} Textual Inversion (\timethod), we adhere to the same training scheme as outlined in the original paper, conducting training for 3000 steps. Subsequently, \ourmethod initializes with the learned tokens from Textual Inversion as a warm-up, followed by an additional 100 steps of continuous updating using both the loss of noise reconstruction and the loss of regularization.
We configure the learning rate as 5e-4 and utilize the AdamW optimizer~\cite{loshchilov2017adamw}. Default hyperparameters are set as $\alpha=1.0$, $\beta=1.0$, and $s=10.0$. Additionally, to expedite training, we extract image features beforehand, ensuring that \ourmethod does not introduce higher time complexity compared to the \timethod approach.
During inference, we adhere to the standard CLIP practice, employing the textual template $T_1$ as \quotes{A photo of a $\pseudoword_k, k \in [1,K]$}. We also investigate the impact of different prompt templates during inference. 
To enhance \ourmethod with CoOp~\cite{zhou2022coop}, we learn a unified context for each dataset with $M=16$ context tokens.
All the experiments are conducted on A40 GPUs.

\minisection{Datasets.}
We evaluated the performance of the few-shot classification in 12 datasets: Oxford-Pets~\cite{parkhi2012pets}, Flowers~\cite{nilsback2008flowers}, Food101~\cite{bossard2014foods}, Aircrafts~\cite{maji2013aircraft}, Stanford-Cars~\cite{krause2013stanfordcars}, CIFAR10~\cite{krizhevsky2009cifar}, STL10~\cite{coates2011stl10}, Caltech101~\cite{fei2004caltech101}, DTD~\cite{cimpoi2014dtd}, EuroSAT~\cite{helber2019eurosat}, UCF101~\cite{soomro2012ucf101} and ImageNet~\cite{deng2009imagenet}.
The datasets chosen for evaluation form a comprehensive benchmark, encompassing a wide array of vision tasks ranging from generic object and scene classification to fine-grained categorization, as well as specialized tasks such as texture recognition and satellite image analysis. Detailed statistics for each dataset are provided in the Appendix.
During the experiments, we \textit{randomly select} $N$-shot ($N=1,2,4,5,8,16$) samples from the training split of each dataset as the $N$-shot train set. The classification performance is then evaluated on the test split.

\minisection{Comparison methods.}
To assess the effectiveness of \ourmethod, we compare with several types of methods:
(i) Four diffusion classification baselines from the Diffusion Classifier (\dicmethod)~\cite{li2023diffusion_classifier}: Synthetic SD Data, SD features, \dicmethod and DM-ZSC~\cite{clark2024dm_zsc} approach.
(ii) CLIP (ResNet-50) and CLIP (ViT-L/14) zero-shot performance as comparisons. 
(iii) prompt learning approaches, including CoOp~\cite{zhou2022coop}, ProGrad~\cite{zhu2023prograd} and PLOT~\cite{chen2022plot} based on RN50~\cite{he2016resnet}.
(iv) \textit{single-concept} Textual Inversion (\timethod)~\cite{textual_inversion} as a baseline for the T2I personalization method.
(v) CLIP-feat is another baseline where we obtain prototype-based classifiers with image features from the CLIP (ViT-L/14) model.
Note that the first three groups of comparison methods are \textit{Semantic-Aware}, which means that the \textit{class names} are required before training.
These last two groups are \textit{Semantic-Agnostic} as \ourmethod.

\minisection{Evaluation metrics.}
We assess the performance across various $N$-shot scenarios ($N=1,2,4,5,8,16$) using classification accuracy on the test set as the primary metric for discriminative performance evaluation. 
Additionally, we calculate the CLIP similarity, a widely used metric in T2I personalization methods~\cite{textual_inversion,gal2023e4t,kumari2022customdiffusion,Cones2023}, which measures the distance between the generated images by \timethod/\ourmethod and the few-shot training samples.
Specifically, for both \timethod and \ourmethod, we randomly generate ten examples with the same prompt \quotes{a photo of a $\pseudoword$} for each category and compute the average CLIP similarity.
This evaluation metric serves to verify the generation performance by assessing whether the learned tokens can successfully \textit{reconstruct} the original concepts. We utilize CLIP (ViT-L/14) for this purpose, consistent with the Stable Diffusion v1.5 setup. All experimental results are averaged over three runs, with standard deviations provided in the Appendix.

\begin{table*}[t]
\setlength{\tabcolsep}{1.5pt}
\caption{
\ourmethod is compared with various approaches. CLIP(RN50) and CLIP(ViT-L) are listed as references. Note that the concepts in CLIP are learned from millions of text-image pairs. We highlight the best performance with \textbf{bold} font and the second with {\underit{underlines}}.
} 
	\centering
	\resizebox{0.97\textwidth}{!}{%
        \begin{tabular}{c| cccccc|c | c c |   ccc | c c c c | c c}
        \toprule
        \multirow{1}{*}{{Method}}  & \multicolumn{6}{c|}{\ourmethod (Ours)} & \tabincell{c}{Ours +\\CoOp} & TI & \tabincell{c}{CLIP\\ViT-L \\ Feat.} & CoOp & ProG & PLOT &{\rotatebox{0}{\makecell{Synth.}}}	& {\rotatebox{0}{{\tabincell{c}{SD \\ Feat.}}}}		& \rotatebox{0}{{\tabincell{c}{\dicmethod}}}	& \tabincell{c}{DM-\\ZSC} &  \rotatebox{0}{{\tabincell{c}{CLIP\\RN50}}}	& \rotatebox{0}{{\tabincell{c}{CLIP\\ViT-L}}}		 \\
        
        \midrule
        \tabincell{c}{Semantic\\Prior} & \multicolumn{9}{c|}{\xmarkg}  & \multicolumn{9}{c}{\cmarkg} \\
        \midrule
        $N$-shot & 1 & 2 & 4 & 5 & 8 & 16 & 16 & 16 & 16 & 16 & 16 & 16&Zero & Full & Zero & Zero& Zero & Zero \\
        \midrule
        
        Ox. Pets & 65.2 &	77.8 &	84.6 &	88.7	& 89.8	&  {{91.7}} & \textbf{94.1} & 50.7 & 73.8   & 87.2  & 89.0 & 87.0 & 31.3 &75.9 & 87.3 & 72.5 & 85.4 & \underit{93.5} \\
        
        Flowers & 80.3 & 87.3 & 91.8 & 93.1 & {{94.8}} & \textbf{95.9} & \underit{95.8} & 40.9 &  73.9 &  94.8 &  94.4 & 94.5 & 22.1 & 70.0 & 66.3 & - & 65.9 & 78.7  \\
        
        Food101 & 53.6 & 68.6 & 77.6 & 80.4 & 82.2 & {{86.0}} & \underit{88.2} & 56.0 & 78.7 & 77.1 &  78.4 & 74.5 &12.6 & 73.0 & 77.7 & 71.6 & 81.1 & \textbf{92.9}  \\
        
        Aircrafts & 24.9 &  32.2 & 39.0 & 40.0 & {{45.5}} & \underit{49.2} & \textbf{51.5}&  16.3 & 40.6  & 31.5 &  31.1 &  31.4 &9.4 & 35.2 & 26.4 &  -& 19.3 & 36.1 \\
        
        Cars & 54.9 & 65.8 & 71.5 & 73.1 & {{77.6}} & \underit{79.5} & \textbf{79.9} &  43.3  & 72.1 & 72.8 &  73.5 & 73.6 &- & - & - &  -& 55.8 & 77.3 \\
        
        CIFAR10 & 61.4 & 78.9 & 86.6 & 91.6 & 91.7 & {{93.4}} & \textbf{96.5} & 31.2 & 76.6 & - & - & - &  35.3 & 84.0 & 88.5 & 72.1 & 75.6 & \underit{96.2} \\
        
        STL10 & 61.8 & 91.9 & 94.4 & 97.5 & 98.4 & {{98.7}} & \underit{98.8} & 65.4 & 93.8 & - & - & - &  38.0 & 87.2 & 95.4 & 92.8 & 94.3 & \textbf{99.3} \\
        
        Caltech & 79.8 & 85.4 & 89.3 & 89.9 & 92.4 & \textbf{93.2} & \underit{93.0} & 59.8 & 87.4 & 92.2 &  92.2 &  92.0 &-& -& - & - & 82.1 &  {{92.6}} \\
        
        DTD & 37.3 & 46.5 & 52.4 & 56.9 & {{60.5}}  & \textbf{65.5} & \underit{65.1} & 34.7  & 54.8 & {63.3} &  64.0 & 62.5 &-& -& - &  -& 41.7 & 55.3 \\
        
        EuroSAT & 60.1 & 62.0 & 71.3 & 72.3 & {{72.8}} & {79.2} & \textbf{85.2} & 13.6 & 69.2 & 82.2 &  \underit{83.7} & {83.6} &- & - & - &  -& 41.1 & 59.9 \\
        
        UCF101 & 55.0 & 58.2 & 61.6 & 61.9 & 69.4 & {77.2} & \textbf{77.5} & 33.0 & 62.4  & {76.9} &  \underit{77.3} & 76.9 &- & - &- &  -& 63.6 & {{76.2}} \\
        
        ImageNet & 41.3 & 53.6 & 63.9 &	 66.4 & 71.2 & \underit{74.8} & 67.9 & 35.0 & 58.9 & 63.0 &  63.5 & 61.9&18.9 & 56.6 & 61.4 & 61.9 & 59.6 & \textbf{75.3}\\
        \midrule
        Average & 56.3	& 67.4	& 73.7	& 76.0	& 78.9	& \underit{82.0}	& \textbf{82.8}	& 40.0	& 70.2 & - & - & - & - & - & - &  - & 63.8& 	77.8 \\
        \bottomrule
        \end{tabular}
	}
% \vspace{-6mm}
\label{tab:eval_metric}
\end{table*}

\begin{table}[t]
\caption{
Comparison between \ourmethod and \timethod in image generation across various datasets by computing the CLIP similarity (\%) between the training few-shot samples and the generated images.
% of both methods. Superior scores are highlighted in \textbf{bold}.
} 
% \vspace{-2mm}
	\centering
	\resizebox{1.05\columnwidth}{!}{%
        \begin{tabular}{c| c|c|c|c|c|c }
        \toprule
         & \multicolumn{6}{c}{CLIP-Similarity (\ourmethod  / \timethod)} \\
        \midrule
        $N$-shot & 1 &2 &4 &5  & 8 &16 \\
        \midrule
        Pets &  \textbf{82.3}/81.8 & \textbf{81.3}/80.6 &  \textbf{81.2}/80.0 &   \textbf{81.0}/80.6 &  \textbf{81.5}/80.3 & \textbf{81.1}/80.5 \\
        Flowers & \textbf{81.1}/81.0 & 81.8/\textbf{81.9} & 81.7/\textbf{81.8} &  {81.9}/\textbf{82.3} & 82.7/\textbf{83.0} & 82.8/\textbf{83.2} \\
        Food101 &  \textbf{77.3}/75.8 & \textbf{77.9}/77.3  & 77.7/\textbf{78.6} & 77.5/\textbf{78.3} & 77.4/\textbf{78.7} & 77.3/\textbf{78.8} \\
        % Aircrafts & \textbf{77.0}/76.4 & \textbf{76.1}/76.0 &  76.2/\textbf{76.5} & 76.4/\textbf{76.5} & 76.3/\textbf{76.8} & 76.2/\textbf{76.8} \\
        Cars & \textbf{78.9}/77.9 & \textbf{78.7}/78.5 & {78.6}/\textbf{78.8} & \textbf{78.7}/78.3 & \textbf{78.5}/78.4 & \textbf{78.7}/78.5 \\
        \bottomrule
        \end{tabular}
	}
% \vspace{-5mm}
\label{tab:clip_img_sim}
\end{table}

\begin{table*}[t]
\vspace{-2mm}
\caption{Ablation study on the Oxford-Pets dataset by varying the $s$, $\beta$, $\alpha$, inference step, template $TPL$ and also the SD versions.
} 
\vspace{-3mm}
	\centering
	\resizebox{0.799\textwidth}{!}{%
        \begin{tabular}{ cc | cccccc | ccc|c }
        \toprule
        \multicolumn{2}{c|}{HyperP} & \multicolumn{6}{c|}{\ourmethod (step=100, TPL=$T_1$, $\alpha=1.0$)} & \multicolumn{3}{c|}{HyperP} &\multicolumn{1}{c}{$s=10.0, \beta=1.0$}  \\
        \midrule
        \multirow{2}{*}{{$s$}} & \multirow{2}{*}{{$\beta$}} & \multicolumn{6}{c|}{$N$-shot} & \multirow{2}{*}{{step}} & \multirow{2}{*}{{TPL}}  & \multirow{2}{*}{$\alpha$} & $N$-shot \\
         &   & 1 & 2 & 4 & 5 & 8 & 16  &  &  & & 5 \\
        \midrule
        3.0 & 1.0   & 62.2 &	76.3 &	84.0 &	86.4	& 88.2	&  90.7  & 50/75 & $T_1$ & 1.0 & 88.6 / 88.8 \\
        \textit{10.0} & \textit{1.0} &  \textbf{65.2} &	\textbf{77.8} &\textbf{	84.6} &	\textbf{88.7}	& \textbf{89.8}	&  \textbf{91.7} & 100 & $T_2/T_3$ & 1.0& 89.0  \\
        30.0 & 1.0 &  61.2 &	74.1 &	82.0 &	85.6	& 87.6	&  88.2  & 100 & $T_1$ & 0.0 & 83.7  \\
        \cline{9-12}
        10.0 & 0.1 &  49.5 &	63.8 &	74.9 &	82.8	& 84.8	&  85.4  & \multicolumn{3}{c|}{SD v1.4} & 88.5   \\
        10.0 & 10.0   & 63.8 &	76.4 &	84.2 &	87.6	& 88.8	&  91.1  & \multicolumn{3}{c|}{SD v2.0} & 88.3 \\
        \bottomrule
        \end{tabular}
	}
\vspace{-3mm}
\label{tab:ablation}
\end{table*}

\begin{figure*}[t]
\centering
\includegraphics[width=0.999\textwidth]{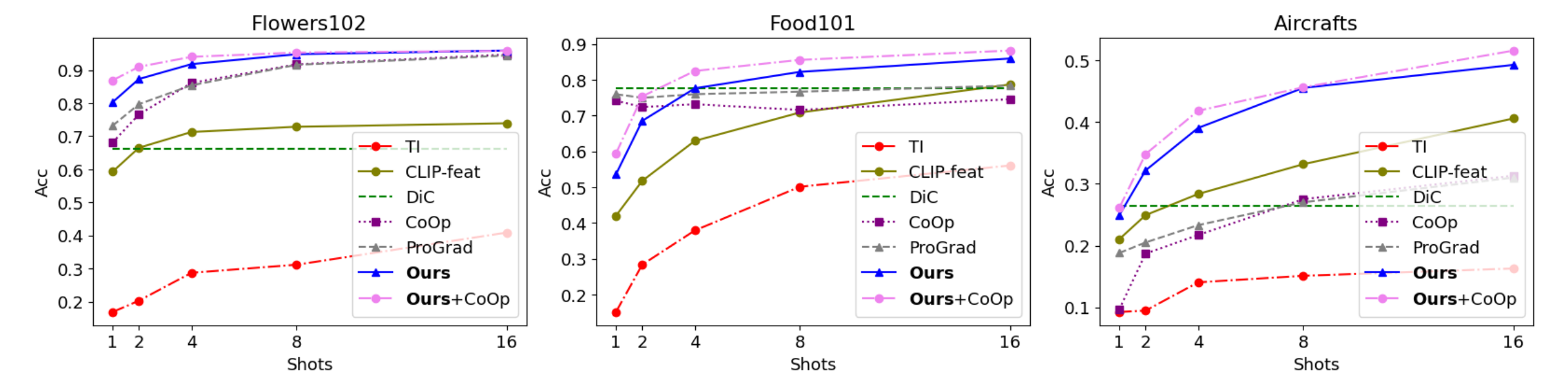}
\vspace{-5mm}
\caption{
\ourmethod is compared with the Textual Inversion (\timethod), the CLIP-feat baseline, Diffusion Classifier (\dicmethod) and prompt learning methods (CoOp, ProGrad) by computing classification accuracies. We vary the $N$-shot ($N=1,2,4,8,16$) numbers to draw the trend plots.
}
\vspace{-1mm}
\label{fig:vsTI}
\end{figure*}

\begin{figure*}[t]
\centering
\includegraphics[width=0.857\textwidth]{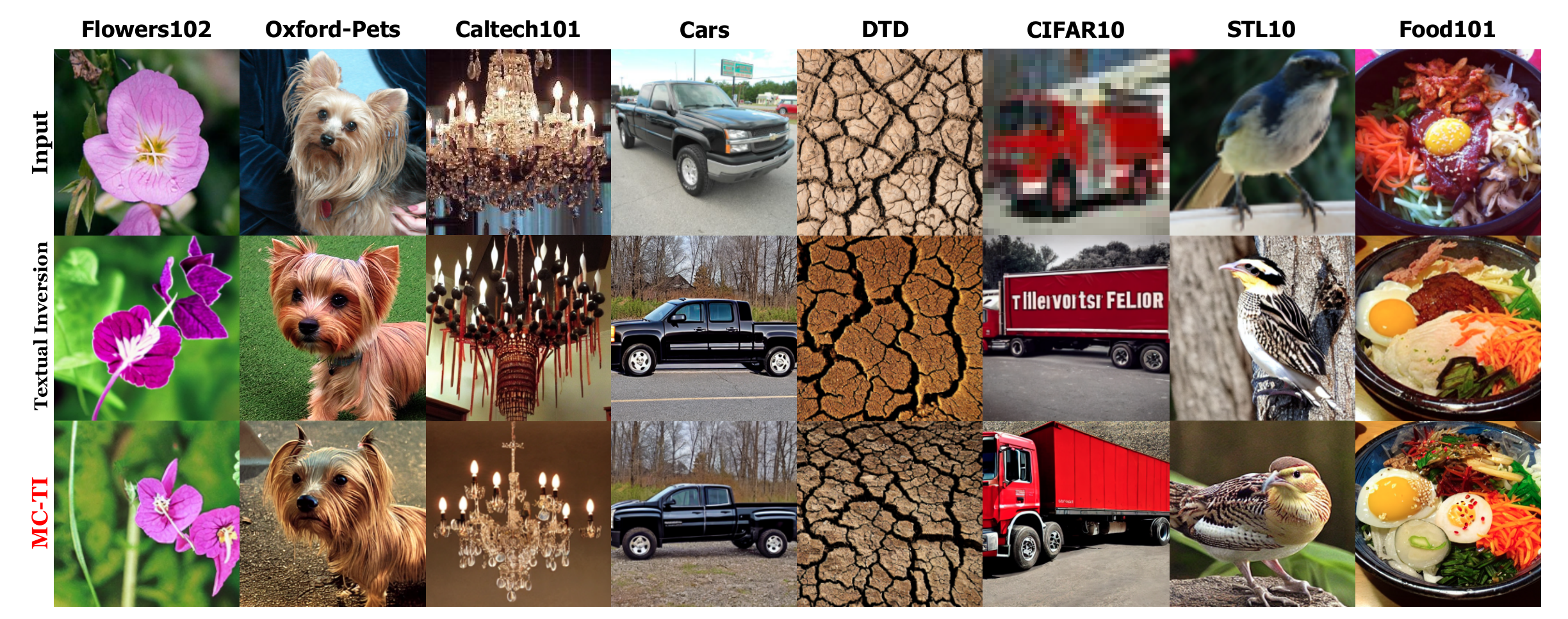}
\vspace{-4mm}
\caption{
\ourmethod does not compromise the generation performance of \timethod as both approaches produce objects similar to the inputs. The illustrated examples are from the 5-shot setup.
}
\vspace{-6mm}
\label{fig:gen_comparison}
\end{figure*}

\subsection{Experimental Results}

\minisection{Discriminative performance} with classification accuracies are shown in Table~\ref{tab:eval_metric}, where our method (\ourmethod and \ourmethod+CoOp) are compared with five groups of comparison methods.
Across the datasets examined in \dicmethod, our approach \ourmethod consistently outperforms with no more than five samples. Remarkably, in nine datasets, our method achieves peak performances with 16-shot samples per class, exceeding even the powerful pretrained model CLIP (ViT-L/14). This trend is particularly pronounced in the Flowers and EuroSAT datasets, where we outperform the competition with CLIP (ViT-L) by 1.6\% and 0.2\% respectively, using only one-shot samples. 
In the remaining datasets, including Food101, STL10 and ImageNet, we also rank as the second-best, trailing only behind the robust model CLIP (ViT-L).
Furthermore, \ourmethod always surpasses \dicmethod and DM-ZSC even with no more than 5-shot examples. Moreover, the CoOp~\cite{zhou2022coop} context optimization can further enhance the performance of  \ourmethod in most datasets.

The tendency curves evaluated over three datasets, depicting the performance with increasing $N$-shot, are illustrated in Fig.~\ref{fig:vsTI}. Here we compare our method with semantic-agnostic baselines (\timethod, CLIP-feat) and prompting approaches.
By comparison, our method \ourmethod consistently outperforms them by no more than 4 shots, especially over the Aircraft dataset. And the CoOp prompting can always improve \ourmethod under various $N$-shot setups.

\minisection{Generative performance} is shown in Table~\ref{tab:clip_img_sim}. 
In this paper, we aim to improve the discriminative capability of the newly learned tokens. 
However, here we verify that our adaptation does not negatively impact the generations.
The comparison reveals that the additional regularization term $\mathcal{L}_{reg}$ does not significantly affect the  generation quality and may even lead to marginal gains.
This conclusion is further supported by the generated samples in Fig.~\ref{fig:gen_comparison}. Considering both discriminative and generative performance, \ourmethod effectively achieves classification objectives without compromising its generative capabilities. 

\minisection{Time complexity.} 
Our method exhibits similar time complexity as \timethod. For learning each concept, the single-concpet TI consumes 12 minute and \ourmethod with 12m24s. 
Note that, both TI and \ourmethod support parallel training for multiple new concepts and do not incur additional time cost in classification phase (as CLIP model).
\dicmethod is achieving classification based on the diffusion process, which incurs a significant sampling time. 
\dicmethod ranges from 18s/image (Oxford-Pets) to 1000s/image (ImageNet), which leads to an unsatisfactory time complexity for real-time classification tasks.

\subsection{Ablation Study}
\vspace{-2mm}
Ablation studies are conducted on the Oxford-Pets dataset, with summarized results in Table~\ref{tab:ablation}.
The scale factor $s$ is varied from 3.0 to 30, revealing that lower values of $s$ correspond to less performance degradation. Similarly, the hyperparameter $\beta$ ranges from 0.1 to 10.0, with \ourmethod showing a preference for higher values of $\beta$, although excessively high values can lead to a decrease in the generation performance.
We also investigate the influence of varying the training steps of our method, observing minimal impact on performance, indicating that \ourmethod converges robustly with only 100 steps.
Lastly, we explore changes in textual templates during inference. By default, we use the same zero-shot classification template as in CLIP~\cite{radford2021clip}, denoted as $T_1$=\quotes{a photo of a {$\pseudoword$}}. Additionally, we experiment with $T_2$=\quotes{a photo of the nice {$\pseudoword$}} and $T_3$=\quotes{a cropped photo of the {$\pseudoword$}}. The results demonstrate the robustness of our method to the templates.
Subsequently, we set $\alpha=0.0$ and update the tokens solely based on the loss of regularization $\mathcal{L}_{reg}$, without initialization from \timethod and without applying the loss of noise reconstruction $\mathcal{L}_{mse}$. In this configuration, the reconstruction quality cannot be guaranteed, and we observe a performance drop of nearly 5\%. This indicates that the generation quality contributes to learning meaningful semantic information, thereby influencing performance. Finally, we also assess \ourmethod using different backbones, namely SD v1.4 and SD v2.0. The classification performances do not exhibit significant changes, as also observed in  \dicmethod~\cite{li2023diffusion_classifier}.

\section{Conclusion}
\vspace{-2mm}
Existing prompt learning methods face challenges while there is no \textit{semantic-aware} prior knowledge of the few samples, where in most cases the \textit{class names} are unknown.
To learn a textual token to represent the class name, \textit{single-concept} T2I adaptation methods excel at learning new concepts from minimal image data. 
However, they often neglect the \textit{discriminative} potential of newly acquired tokens. This study delves into \textit{single-concept Textual Inversion} as a representative of the T2I adaptation.
Our investigation uncovers the dual nature of \textit{multiple concept} tokens, possessing both generative and discriminative capabilities. However, token updates may lack directionality without proper constraints. To mitigate this issue, we introduce an additional regularization term. Our \textit{Multi-Class Textual Inversion} method, named \ourmethod, achieves a robust \textit{Semantic-Agnostic Classification} by incorporating discriminative regularization while retaining the generative prowess of modifier tokens. Extensive evaluations of diverse datasets consistently show superior results in both classification and generation performance.

\minisection{Acknowledgements.}
This work is funded by Grants TED2021-132513B-I00 and PID2022-143257NB-I00 funded by MCIN/AEI/10.13039/501100011033, by the European Union NextGenerationEU/PRTR and by ERDF A Way of Making Europa, the Departament de Recerca i Universitats from Generalitat de Catalunya with reference 2021SGR01499, and the Generalitat de Catalunya CERCA.

% \clearpage

%%%%%%%%% REFERENCES
{\small
\bibliographystyle{ieee_fullname}
\bibliography{longstrings,mybib}

\begin{thebibliography}{10}\itemsep=-1pt

\bibitem{alayrac2022flamingo}
Jean-Baptiste Alayrac, Jeff Donahue, Pauline Luc, Antoine Miech, Iain Barr, Yana Hasson, Karel Lenc, Arthur Mensch, Katherine Millican, Malcolm Reynolds, et~al.
\newblock Flamingo: a visual language model for few-shot learning.
\newblock {\em Advances in neural information processing systems}, 35:23716--23736, 2022.

\bibitem{bossard2014foods}
Lukas Bossard, Matthieu Guillaumin, and Luc Van~Gool.
\newblock Food-101--mining discriminative components with random forests.
\newblock In {\em European Conference on Computer Vision}, pages 446--461. Springer, 2014.

\bibitem{butt2025colorpeel}
Muhammad~Atif Butt, Kai Wang, Javier Vazquez-Corral, and Joost van~de Weijer.
\newblock Colorpeel: Color prompt learning with diffusion models via color and shape disentanglement.
\newblock In {\em European Conference on Computer Vision}, pages 456--472. Springer, 2025.

\bibitem{chen2022plot}
Guangyi Chen, Weiran Yao, Xiangchen Song, Xinyue Li, Yongming Rao, and Kun Zhang.
\newblock Prompt learning with optimal transport for vision-language models.
\newblock {\em International Conference on Learning Representations}, 2023.

\bibitem{cimpoi2014dtd}
Mircea Cimpoi, Subhransu Maji, Iasonas Kokkinos, Sammy Mohamed, and Andrea Vedaldi.
\newblock Describing textures in the wild.
\newblock In {\em Proceedings of the IEEE conference on computer vision and pattern recognition}, pages 3606--3613, 2014.

\bibitem{clark2024dm_zsc}
Kevin Clark and Priyank Jaini.
\newblock Text-to-image diffusion models are zero shot classifiers.
\newblock {\em Advances in Neural Information Processing Systems}, 36, 2024.

\bibitem{coates2011stl10}
Adam Coates, Andrew Ng, and Honglak Lee.
\newblock An analysis of single-layer networks in unsupervised feature learning.
\newblock In {\em Proceedings of the fourteenth international conference on artificial intelligence and statistics}, pages 215--223. JMLR Workshop and Conference Proceedings, 2011.

\bibitem{deng2009imagenet}
Jia Deng, Wei Dong, Richard Socher, Li-Jia Li, Kai Li, and Li Fei-Fei.
\newblock Imagenet: A large-scale hierarchical image database.
\newblock In {\em 2009 IEEE conference on computer vision and pattern recognition}, pages 248--255. Ieee, 2009.

\bibitem{devlin2018bert}
Jacob Devlin, Ming-Wei Chang, Kenton Lee, and Kristina Toutanova.
\newblock Bert: Pre-training of deep bidirectional transformers for language understanding.
\newblock {\em arXiv preprint arXiv:1810.04805}, 2018.

\bibitem{fei2004caltech101}
Li Fei-Fei, Rob Fergus, and Pietro Perona.
\newblock Learning generative visual models from few training examples: An incremental bayesian approach tested on 101 object categories.
\newblock In {\em 2004 conference on computer vision and pattern recognition workshop}, pages 178--178. IEEE, 2004.

\bibitem{textual_inversion}
Rinon Gal, Yuval Alaluf, Yuval Atzmon, Or Patashnik, Amit~H Bermano, Gal Chechik, and Daniel Cohen-Or.
\newblock An image is worth one word: Personalizing text-to-image generation using textual inversion.
\newblock {\em International Conference on Learning Representations}, 2023.

\bibitem{gal2023e4t}
Rinon Gal, Moab Arar, Yuval Atzmon, Amit~H Bermano, Gal Chechik, and Daniel Cohen-Or.
\newblock Designing an encoder for fast personalization of text-to-image models.
\newblock {\em arXiv preprint arXiv:2302.12228}, 2023.

\bibitem{gao2024clip_adapter}
Peng Gao, Shijie Geng, Renrui Zhang, Teli Ma, Rongyao Fang, Yongfeng Zhang, Hongsheng Li, and Yu Qiao.
\newblock Clip-adapter: Better vision-language models with feature adapters.
\newblock {\em International Journal of Computer Vision}, 132(2):581--595, 2024.

\bibitem{gu2024mixofshow}
Yuchao Gu, Xintao Wang, Jay~Zhangjie Wu, Yujun Shi, Yunpeng Chen, Zihan Fan, Wuyou Xiao, Rui Zhao, Shuning Chang, Weijia Wu, et~al.
\newblock Mix-of-show: Decentralized low-rank adaptation for multi-concept customization of diffusion models.
\newblock {\em Advances in Neural Information Processing Systems}, 36, 2024.

\bibitem{hiper2023}
Inhwa Han, Serin Yang, Taesung Kwon, and Jong~Chul Ye.
\newblock Highly personalized text embedding for image manipulation by stable diffusion.
\newblock {\em arXiv preprint arXiv:2303.08767}, 2023.

\bibitem{han2023svdiff}
Ligong Han, Yinxiao Li, Han Zhang, Peyman Milanfar, Dimitris Metaxas, and Feng Yang.
\newblock Svdiff: Compact parameter space for diffusion fine-tuning.
\newblock {\em Proceedings of the International Conference on Computer Vision}, 2023.

\bibitem{he2022masked}
Kaiming He, Xinlei Chen, Saining Xie, Yanghao Li, Piotr Doll{\'a}r, and Ross Girshick.
\newblock Masked autoencoders are scalable vision learners.
\newblock In {\em Proceedings of the IEEE Conference on Computer Vision and Pattern Recognition}, pages 16000--16009, 2022.

\bibitem{he2016resnet}
Kaiming He, Xiangyu Zhang, Shaoqing Ren, and Jian Sun.
\newblock Deep residual learning for image recognition.
\newblock In {\em Proceedings of the IEEE conference on computer vision and pattern recognition}, pages 770--778, 2016.

\bibitem{helber2019eurosat}
Patrick Helber, Benjamin Bischke, Andreas Dengel, and Damian Borth.
\newblock Eurosat: A novel dataset and deep learning benchmark for land use and land cover classification.
\newblock {\em IEEE Journal of Selected Topics in Applied Earth Observations and Remote Sensing}, 12(7):2217--2226, 2019.

\bibitem{hinton2007recognize}
Geoffrey~E Hinton.
\newblock To recognize shapes, first learn to generate images.
\newblock {\em Progress in brain research}, 165:535--547, 2007.

\bibitem{jia2021ALIGN}
Chao Jia, Yinfei Yang, Ye Xia, Yi-Ting Chen, Zarana Parekh, Hieu Pham, Quoc Le, Yun-Hsuan Sung, Zhen Li, and Tom Duerig.
\newblock Scaling up visual and vision-language representation learning with noisy text supervision.
\newblock In {\em International conference on machine learning}, pages 4904--4916. PMLR, 2021.

\bibitem{krause2013stanfordcars}
Jonathan Krause, Michael Stark, Jia Deng, and Li Fei-Fei.
\newblock 3d object representations for fine-grained categorization.
\newblock In {\em Proceedings of the IEEE international conference on computer vision workshops}, pages 554--561, 2013.

\bibitem{krizhevsky2009cifar}
Alex Krizhevsky, Geoffrey Hinton, et~al.
\newblock Learning multiple layers of features from tiny images.
\newblock 2009.

\bibitem{kumari2022customdiffusion}
Nupur Kumari, Bingliang Zhang, Richard Zhang, Eli Shechtman, and Jun-Yan Zhu.
\newblock Multi-concept customization of text-to-image diffusion.
\newblock {\em Proceedings of the IEEE Conference on Computer Vision and Pattern Recognition}, 2023.

\bibitem{li2023diffusion_classifier}
Alexander~C Li, Mihir Prabhudesai, Shivam Duggal, Ellis Brown, and Deepak Pathak.
\newblock Your diffusion model is secretly a zero-shot classifier.
\newblock {\em Proceedings of the International Conference on Computer Vision}, 2023.

\bibitem{li2023stylediffusion}
Senmao Li, Joost van~de Weijer, Taihang Hu, Fahad~Shahbaz Khan, Qibin Hou, Yaxing Wang, and Jian Yang.
\newblock Stylediffusion: Prompt-embedding inversion for text-based editing, 2023.

\bibitem{li2023photomaker}
Zhen Li, Mingdeng Cao, Xintao Wang, Zhongang Qi, Ming-Ming Cheng, and Ying Shan.
\newblock Photomaker: Customizing realistic human photos via stacked id embedding.
\newblock {\em arXiv preprint arXiv:2312.04461}, 2023.

\bibitem{Cones2023}
Zhiheng Liu, Ruili Feng, Kai Zhu, Yifei Zhang, Kecheng Zheng, Yu Liu, Deli Zhao, Jingren Zhou, and Yang Cao.
\newblock Cones: Concept neurons in diffusion models for customized generation.
\newblock {\em International Conference on Machine Learning}, 2023.

\bibitem{liu2015celeba}
Ziwei Liu, Ping Luo, Xiaogang Wang, and Xiaoou Tang.
\newblock Deep learning face attributes in the wild.
\newblock In {\em Proceedings of the International Conference on Computer Vision}, December 2015.

\bibitem{loshchilov2017adamw}
Ilya Loshchilov and Frank Hutter.
\newblock Decoupled weight decay regularization.
\newblock {\em arXiv preprint arXiv:1711.05101}, 2017.

\bibitem{maji2013aircraft}
Subhransu Maji, Esa Rahtu, Juho Kannala, Matthew Blaschko, and Andrea Vedaldi.
\newblock Fine-grained visual classification of aircraft.
\newblock {\em arXiv preprint arXiv:1306.5151}, 2013.

\bibitem{ng2001discriminative}
Andrew Ng and Michael Jordan.
\newblock On discriminative vs. generative classifiers: A comparison of logistic regression and naive bayes.
\newblock {\em Advances in Neural Information Processing Systems}, 14, 2001.

\bibitem{nilsback2008flowers}
Maria-Elena Nilsback and Andrew Zisserman.
\newblock Automated flower classification over a large number of classes.
\newblock In {\em 2008 Sixth Indian conference on computer vision, graphics \& image processing}, pages 722--729. IEEE, 2008.

\bibitem{parkhi2012pets}
Omkar~M Parkhi, Andrea Vedaldi, Andrew Zisserman, and CV Jawahar.
\newblock Cats and dogs.
\newblock In {\em Proceedings of the IEEE Conference on Computer Vision and Pattern Recognition}, pages 3498--3505. IEEE, 2012.

\bibitem{radford2021clip}
Alec Radford, Jong~Wook Kim, Chris Hallacy, Aditya Ramesh, Gabriel Goh, Sandhini Agarwal, Girish Sastry, Amanda Askell, Pamela Mishkin, Jack Clark, et~al.
\newblock Learning transferable visual models from natural language supervision.
\newblock In {\em International conference on machine learning}, pages 8748--8763. PMLR, 2021.

\bibitem{ranzato2011deep}
Marc'Aurelio Ranzato, Joshua Susskind, Volodymyr Mnih, and Geoffrey Hinton.
\newblock On deep generative models with applications to recognition.
\newblock In {\em Proceedings of the IEEE Conference on Computer Vision and Pattern Recognition}, pages 2857--2864. IEEE, 2011.

\bibitem{Rombach_2022_CVPR_stablediffusion}
Robin Rombach, Andreas Blattmann, Dominik Lorenz, Patrick Esser, and Bj\"orn Ommer.
\newblock High-resolution image synthesis with latent diffusion models.
\newblock In {\em Proceedings of the IEEE/CVF Conference on Computer Vision and Pattern Recognition (CVPR)}, pages 10684--10695, 06 2022.

\bibitem{ronneberger2015unet}
Olaf Ronneberger, Philipp Fischer, and Thomas Brox.
\newblock U-net: Convolutional networks for biomedical image segmentation.
\newblock In {\em Medical Image Computing and Computer-Assisted Intervention--MICCAI 2015: 18th International Conference, Munich, Germany, October 5-9, 2015, Proceedings, Part III 18}, pages 234--241. Springer, 2015.

\bibitem{ruiz2022dreambooth}
Nataniel Ruiz, Yuanzhen Li, Varun Jampani, Yael Pritch, Michael Rubinstein, and Kfir Aberman.
\newblock Dreambooth: Fine tuning text-to-image diffusion models for subject-driven generation.
\newblock {\em Proceedings of the IEEE Conference on Computer Vision and Pattern Recognition}, 2023.

\bibitem{soomro2012ucf101}
Khurram Soomro, Amir~Roshan Zamir, and Mubarak Shah.
\newblock Ucf101: A dataset of 101 human actions classes from videos in the wild.
\newblock {\em arXiv preprint arXiv:1212.0402}, 2012.

\bibitem{kai2023DPL}
Kai Wang, Fei Yang, Shiqi Yang, Muhammad~Atif Butt, and Joost van~de Weijer.
\newblock Dynamic prompt learning: Addressing cross-attention leakage for text-based image editing.
\newblock {\em Advances in Neural Information Processing Systems}, 2023.

\bibitem{yao2023kgcoop}
Hantao Yao, Rui Zhang, and Changsheng Xu.
\newblock Visual-language prompt tuning with knowledge-guided context optimization.
\newblock In {\em Proceedings of the IEEE Conference on Computer Vision and Pattern Recognition}, pages 6757--6767, 2023.

\bibitem{zhou2022coop}
Kaiyang Zhou, Jingkang Yang, Chen~Change Loy, and Ziwei Liu.
\newblock Learning to prompt for vision-language models.
\newblock {\em International Journal of Computer Vision}, 130(9):2337--2348, 2022.

\bibitem{zhu2023prograd}
Beier Zhu, Yulei Niu, Yucheng Han, Yue Wu, and Hanwang Zhang.
\newblock Prompt-aligned gradient for prompt tuning.
\newblock In {\em Proceedings of the International Conference on Computer Vision}, pages 15659--15669, 2023.

\end{thebibliography}
}

\clearpage

\appendix
\section*{Supplementary Material}

\section{Boarder Impacts}
The utilization of personalized text-to-image (T2I) models holds promise for a diverse array of applications in various domains. Our model seeks to enhance the dual functionality of the updated tokens in these models. However, it is imperative to acknowledge potential risks, including the dissemination of misinformation, potential misuse, and the introduction of biases. Ethical considerations and broader impacts require a thorough examination to ensure responsible utilization of these models and their capabilities.

\section{Limitations}
One limitation of our approach is its dependence on the Textual Inversion technique, neglecting the thorough exploration into methods that fine-tune diffusion backbones. Moreover, our method necessitates the features of all few-shot samples during training, which might not be available in scenarios where such information is not provided in advance. Lastly, \ourmethod may introduce increased time complexity in datasets featuring thousands of categories.

\section{Experiments}
\minisection{Dataset details.}
The detailed statistics of all the datasets, are shown in Table~\ref{tab:dataset_stat}. The initialization tokens for \timethod~\cite{textual_inversion} and \ourmethod are also detailed in the table. 

\minisection{Mean and Standard deviations.}
The experimental mean and standard deviation values are provided in Table~\ref{tab:eval_meanstd}. These results further confirm the robustness of \ourmethod to randomness.

\minisection{Generative performance.} The full table for generation quality comparison with CLIP-Similarity is shown in Table~\ref{tab:clip_img_sim_supp}, which further demonstrates that \ourmethod is not damaging the generation capability.

\begin{figure*}[t]
\centering
\includegraphics[width=0.79\textwidth]{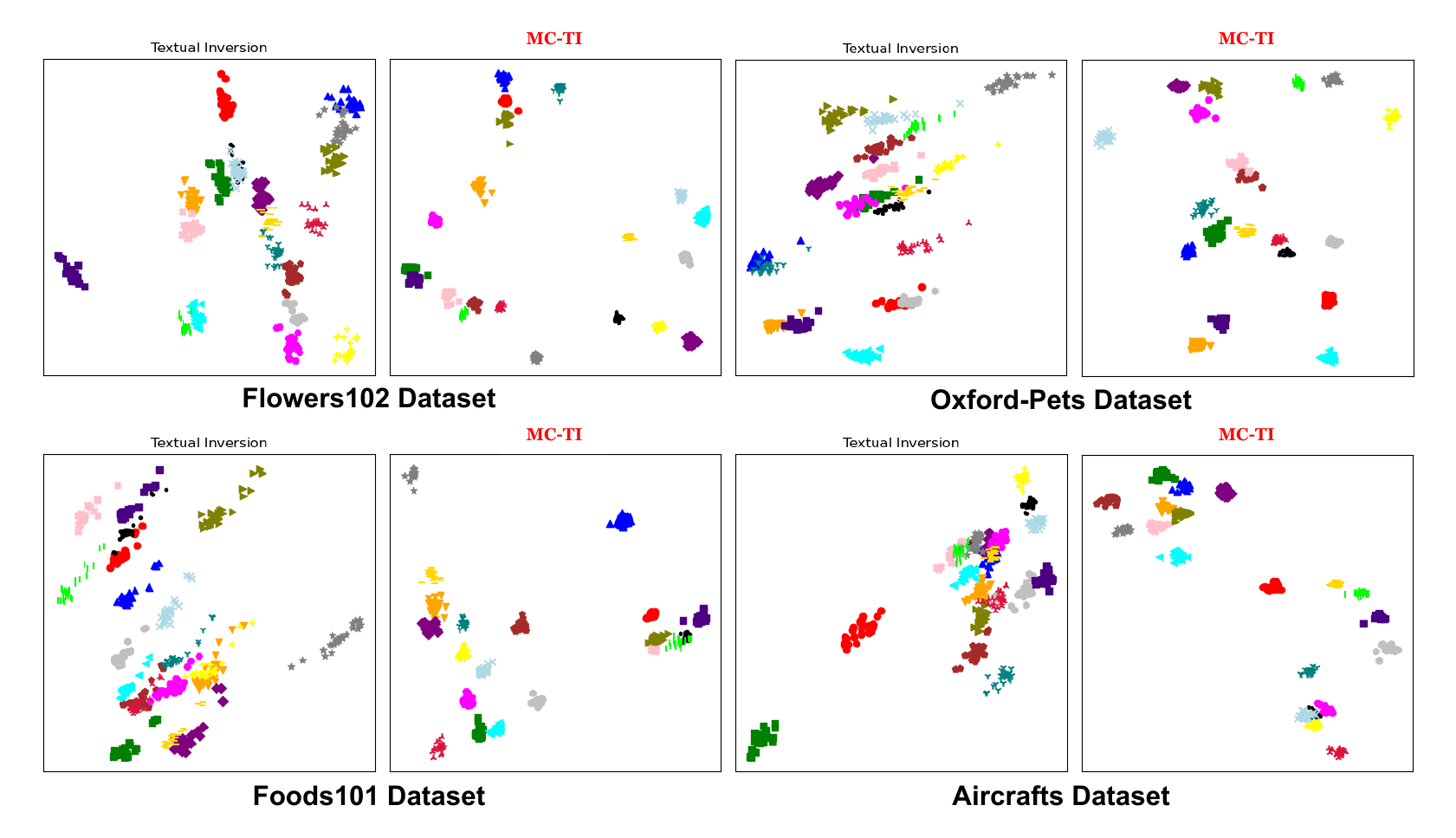}
% \vspace{-5mm}
\caption{
To visualize the textual prompts features, we took the 5-shot conceptual tokens learned by Textual Inversion and \ourmethod, respectively. By applying 27 types of various prompt templates, we visualize the PCA components in 2-D maps. 
}
% \vspace{-5mm}
\label{fig:text_feature_supp}
\end{figure*}

\minisection{Additional textual feature visualization.}
Additionally, we visualize the textual feature changes from \timethod to \ourmethod on four datasets, as depicted in Fig.~\ref{fig:text_feature_supp}. We extract the textual features of 27 prompt templates from 5-shot learning schemes.
As observed, \ourmethod enhances the clustering of textual characteristics by enforcing discriminative regularization terms, resulting in improved classification performances.

\begin{figure*}[t]
\centering
\includegraphics[width=0.79\textwidth]{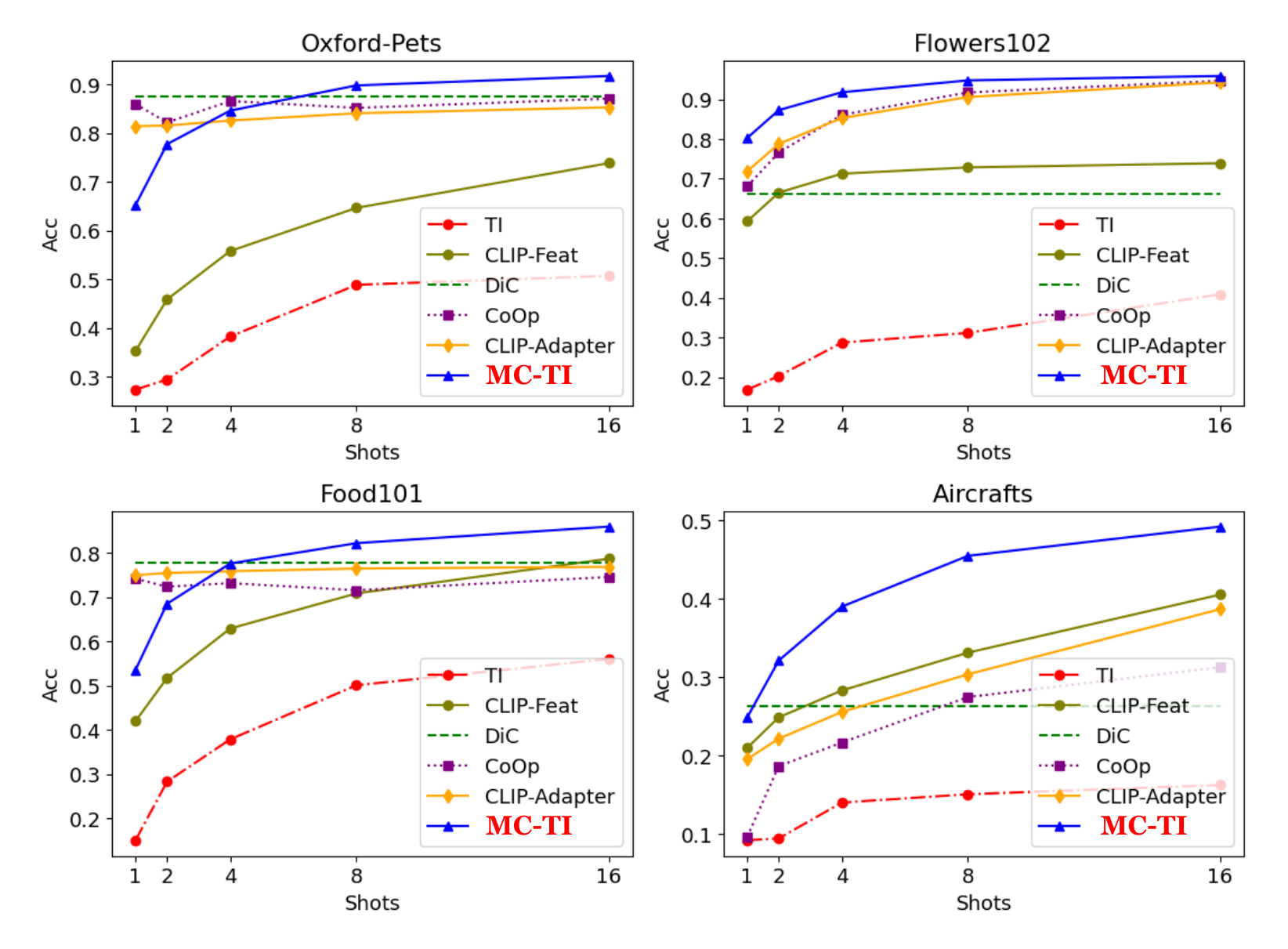}
% \vspace{-5mm}
\caption{
\ourmethod is compared with the Textual Inversion (\timethod)~\cite{textual_inversion}, the CLIP-feat baseline, Diffusion Classifier (\dicmethod)~\cite{li2023diffusion_classifier} and two prompt tuning methods (CoOp~\cite{zhou2022coop} and CLIP-Adapter~\cite{gao2024clip_adapter}) over four fine-grained datasets by computing classification accuracies. We vary the $N$-shot ($N=1,2,4,8,16$) numbers to draw the trend plots.
}
% \vspace{-5mm}
\label{fig:vs_ti_supp}
\end{figure*}

\minisection{Extended comparison with prompt tuning methods.}
In Fig.\ref{fig:vs_ti_supp}, we provide a comparison of our method, \ourmethod, with two classical prompt tuning methods, namely CoOp~\cite{zhou2022coop} and CLIP-Adapter~\cite{gao2024clip_adapter}. It's important to note that these prompt tuning methods primarily operate on the CLIP (ResNet-50) backbones. This comparison serves to further demonstrate that our method is comparable to current prompt tuning approaches.

\clearpage

\begin{table*}[t]
\caption{
The detailed statistics of datasets used in experiments. 
} 
	\centering
	\resizebox{0.93\textwidth}{!}{%
        \begin{tabular}{cccccc}
        \toprule
        Dataset & classes & \tabincell{c}{Train \\ size} & \tabincell{c}{Test \\ size} & Task &   \tabincell{c}{Initialization \\ token}  \\
        \midrule
        Oxford-Pets~\cite{parkhi2012pets} & 37 & 2944 & 3669 & Fine-grained pet recognition & \quotes{pet}  \\
        Flowers~\cite{nilsback2008flowers} & 102 & 4093 & 2463 & Fine-grained flowers recognition & \quotes{flower}\\
        Food101~\cite{bossard2014foods} & 101 & 50500 & 30300 & Fine-grained food recognition & \quotes{food}  \\
        Aircrafts~\cite{maji2013aircraft} & 100 & 3334 & 3333 & Fine-grained aircraft recognition & \quotes{aircraft}\\
        Stanford-Cars~\cite{krause2013stanfordcars} & 196 & 6509 & 8041 & Fine-grained car recognition & \quotes{car}  \\
        CIFAR10~\cite{krizhevsky2009cifar} & 10 & 50000 & 10000 & Generic object recognition & \quotes{object}\\
        STL10~\cite{coates2011stl10} &  10 & 1000 & 8000 & Generic object recognition & \quotes{object} \\
        Caltech101~\cite{fei2004caltech101} &  102 & 4128 & 2465 & Generic object recognition & \quotes{object} \\
        DTD~\cite{cimpoi2014dtd} & 47 & 2820 & 1692 & texture recognition & \quotes{texture} \\
        EuroSAT~\cite{helber2019eurosat} & 10 & 13500 & 8100 & Satellite image recognition & \quotes{object} \\
        UCF101~\cite{soomro2012ucf101} & 101 & 7639 & 3783 & Action recognition & \quotes{action} \\
        \midrule
        ImageNet-1000~\cite{deng2009imagenet} & 1000 & 1.28M & 50000 &  Generic object recognition & \quotes{object}  \\
        \bottomrule
        \end{tabular}
}
\label{tab:dataset_stat}
\end{table*}

\begin{table*}[tb]
\caption{
We present the mean $\pm$ standard deviation of the evaluation accuracies (\%) achieved by \ourmethod on four fine-grained datasets. This analysis demonstrates that randomization does not significantly influence the performance.
} 
% \vspace{-1mm}
	\centering
	\resizebox{0.88\textwidth}{!}{%
        \begin{tabular}{c| c|c|c|c|c|c }
        \toprule
        \multirow{1}{*}{{Method}}  & \multicolumn{6}{c}{\ourmethod (Ours)} 	 \\
        \midrule
        $N$-shot & 1 & 2 & 4 & 5 & 8 & 16  \\
        \midrule
        Oxford-Pets & 65.2$\pm$0.4 &	77.8$\pm$0.5 &	84.6$\pm$1.5 &	88.7$\pm$0.7	& 89.8$\pm$1.8	& 91.7$\pm$0.5 \\
        Flowers & 80.3$\pm$0.7 & 87.3$\pm$0.4 & 91.8$\pm$0.3 & 93.1$\pm$0.9 & 94.8$\pm$0.8 &  95.9$\pm$0.1 \\
        Food101 & 53.6$\pm$0.9 & 68.6$\pm$1.3 & 77.6$\pm$1.4 & 80.4$\pm$0.7 & 82.2$\pm$1.9 &   86.0$\pm$1.7 \\
        Aircrafts & 24.9$\pm$1.9 &  32.2$\pm$2.1 & 39.0$\pm$1.8 & 40.0$\pm$1.0 &   45.5$\pm$0.8 &  49.2$\pm$2.2 \\
        \bottomrule
        \end{tabular}
}
\label{tab:eval_meanstd}
\end{table*}

\begin{table*}[t]
\caption{
Comparison between \ourmethod and \timethod in image generation across eleven datasets by computing the CLIP similarity (\%) between the training few-shot samples and the generated images of both methods. Superior scores are highlighted in \textbf{bold}.
} 
% \vspace{-2mm}
	\centering
	\resizebox{0.88\textwidth}{!}{%
        \begin{tabular}{c| c|c|c|c|c|c }
        \toprule
         & \multicolumn{6}{c}{CLIP-Similarity (\ourmethod / \timethod)} \\
        \midrule
        $N$-shot & 1 &2 &4 &5  & 8 &16 \\
        \midrule
        Oxford Pets &  \textbf{82.3} / 81.8 & \textbf{81.3} / 80.6 &  \textbf{81.2} / 80.0 &   \textbf{81.0} / 80.6 &  \textbf{81.5} / 80.3 & \textbf{81.1} / 80.5 \\
        Flowers & \textbf{81.1} / 81.0 & 81.8 / \textbf{81.9} & 81.7 / \textbf{81.8} &  {81.9} / \textbf{82.3} & 82.7 / \textbf{83.0} & 82.8 / \textbf{83.2} \\
        Food101 &  \textbf{77.3} / 75.8 & \textbf{77.9} / 77.3  & 77.7 / \textbf{78.6} & 77.5 / \textbf{78.3} & 77.4 / \textbf{78.7} & 77.3 / \textbf{78.8} \\
        Aircrafts & \textbf{77.0} / 76.4 & \textbf{76.1} / 76.0 &  76.2 / \textbf{76.5} & 76.4 / \textbf{76.5} & 76.3 / \textbf{76.8} & 76.2 / \textbf{76.8} \\
        Stanford Cars & \textbf{78.9} / 77.9 & \textbf{78.7} / 78.5 & {78.6} / \textbf{78.8} & \textbf{78.7} / 78.3 & \textbf{78.5} / 78.4 & \textbf{78.7} / 78.5 \\
        CIFAR10 & \textbf{65.4} / 62.2 & \textbf{65.8} / 63.2 & \textbf{67.2} / 64.1 & \textbf{65.7} / 64.5 &  \textbf{65.4} / 64.7 & \textbf{66.3} / 64.8 \\
        STL10 & \textbf{75.0} / 73.2 & \textbf{74.7} / 72.5 & \textbf{70.1} / 65.6 & \textbf{70.6} / 68.2  & \textbf{71.1} / 68.4 & \textbf{69.3} / 67.8 \\
        Caltech101 & \textbf{75.8} / 73.7 & \textbf{76.4} / 73.9 & \textbf{75.2} / 74.6 & \textbf{75.5} / 74.2 & \textbf{75.1} / 73.9 & \textbf{74.7} / 73.7  \\
        DTD & \textbf{78.5} / 77.7 & \textbf{73.8} / 73.2 & \textbf{72.8} / 72.5 & \textbf{72.6} / 72.2 & \textbf{73.3} / 72.8 & \textbf{72.8} / 72.0 \\
        EuroSAT & 57.6 / \textbf{58.7} & \textbf{60.7} / 60.1 & 57.9 / \textbf{58.8} & \textbf{60.9} / 60.7  & 59.1 / \textbf{59.6} & 58.6 / \textbf{58.7}   \\
        UCF101 & 61.3 / \textbf{61.8} & 62.1 / \textbf{62.9} & \textbf{63.3} / 63.2 & 62.8 / \textbf{62.9} & 62.0 / \textbf{62.7} & \textbf{63.2} / 62.4 \\
        \bottomrule
        \end{tabular}
	}
% \vspace{-2mm}
\label{tab:clip_img_sim_supp}
\end{table*}

\end{document}